\documentclass{bmvc2k}

\usepackage{booktabs}   
\PassOptionsToPackage{table}{xcolor}
\usepackage{xcolor}
\definecolor{lightblue}{RGB}{220,230,250}
\usepackage{tikz}
\usepackage{colortbl}
\usepackage{amssymb}

\title{Fast Self-Supervised depth and mask aware Association for Multi-Object Tracking}

\addauthor{Milad Khanchi}{milad.khanchi@concordia.ca}{1}
\addauthor{Maria Amer}{amer@ece.concordia.ca}{1}
\addauthor{Charalambos Poullis}{charalambos@poullis.org}{1}

\addinstitution{
 Concordia University\\
 Montreal, QC, Canada
}

\runninghead{Khanchi, Amer, Poullis}{Self-sup. Depth-Mask Aware Assoc. for MOT}

\begin{document}

\maketitle

\begin{abstract}
Multi-object tracking (MOT) methods often rely on Intersection-over-Union (IoU) for association. However, this becomes unreliable when objects are similar or occluded. Also, computing IoU for segmentation masks is computationally expensive. In this work, we use segmentation masks to capture object shapes, but we do not compute segmentation IoU. Instead, we fuse depth and mask features and pass them through a compact encoder trained self-supervised. This encoder produces stable object representations, which we use as an additional similarity cue alongside bounding box IoU and re-identification features for matching. We obtain depth maps from a zero-shot depth estimator and object masks from a promptable visual segmentation model to obtain fine-grained spatial cues. Our MOT method is the first to use the self-supervised encoder to refine segmentation masks without computing masks IoU. MOT can be divided into joint detection-ReID (JDR) and tracking-by-detection (TBD) models. The latter are computationally more efficient. Experiments of our TBD method on challenging benchmarks with non-linear motion, occlusion, and crowded scenes, such as SportsMOT and DanceTrack, show that our method outperforms the TBD state-of-the-art on most metrics, while achieving competitive performance on simpler benchmarks with linear motion, such as MOT17. The code is available at \href{https://github.com/Milad-Khanchi/SelfTrEncMOT}{https://github.com/Milad-Khanchi/SelfTrEncMOT}.
\end{abstract}

\section{Introduction}
\label{sec:intro}

Multi-Object Tracking (MOT)~\cite{li2025self, lv2024diffmot, shim2024confidence} aims to detect and maintain object identities across video frames. Despite notable advances, existing approaches still struggle under conditions of occlusion, appearance similarity, and rapid motion. These challenges are amplified when 2D cues, such as bounding box overlap (IoU) and appearance re-identification (Re-ID), are the sole drivers of association. For instance, two pedestrians walking in parallel but at different depths may appear indistinguishable in 2D, leading to frequent ID switches.

To address these limitations, we propose a combined depth and segmentation aware method that supplements traditional 2D cues with pixel-aligned geometric reasoning. Specifically, we combine zero-shot monocular depth estimation with promptable segmentation masks to extract fine-grained spatial features. The combined depth-segmentation embeddings are processed by a lightweight, self-supervised encoder that enhances temporal consistency and reduces noise caused by artifacts from segmentation or depth map. The resulting features serve as an additional matching score during data association, complementing motion and appearance 2D cues.

Unlike prior works~\cite{quach2024depth, liu2022det, wang2024robust} that use depth or segmentation as auxiliary inputs, our approach introduces depth-segmentation cues as an explicit similarity matrix for matching. This design enables robust identity preservation in occluded and visually ambiguous scenes.

We validate our method on challenging MOT benchmarks DanceTrack and SportsMOT, which feature crowded and occluded objects with complex non-linear motion, and demonstrate consistent improvements in association-based metrics such as HOTA, IDF1, and AssA. Our method achieves competitive performance on benchmarks with simpler, mostly linear motion, such as MOT17. Our contributions are: 1) We design a self-supervised encoder to enhance depth-segmentation features' temporal stability and discriminability. 2) Our MOT method is the first to use the self-supervised encoder to refine segmentation masks and integrate them into the matching score without computing mask IoU. 3) We achieve competitive performance across various tracking scenarios, especially in occluded scenes.

\section{Prior Works}
\label{sec:RELATEDWORKS}

\noindent \textbf{Joint Detection-ReID methods (JDR):} They unify detection and tracking in a single forward pass~\cite{zhang2021fairmot, xu2022transcenter}. FairMOT~\cite{zhang2021fairmot} pioneered a dual-branch approach combining anchor-free detection with appearance embeddings. TransCenter~\cite{xu2022transcenter} extends this with deformable attention, enabling improved occlusion handling. Recent models in this category go beyond earlier dual-branch architectures by integrating attention mechanisms or spatial alignment strategies. AFMTrack~\cite{bui2024afmtrack} enhances identity preservation by introducing an attention-based feature matching network, allowing robust association even in dense scenes. DilateTracker~\cite{wu2024dilatetracker} integrates dilated attention modules into the joint detection-ReID framework, significantly boosting identity recall under occlusion. 


\noindent \textbf{Tracking by Detection methods (TBD):} The tracking-by-detection paradigm remains dominant in recent literature, where objects are first localized in each frame and then associated temporally~\cite{vaquero2024lost, lv2024diffmot, cao2023observation, maggiolino2023deep, zhu2018online, meinhardt2022trackformer, zhang2023motrv2}. The effectiveness of these methods relies on detection quality and the design of robust association. Depending on their temporal processing technique, these methods can be categorized into sequence-level and frame-wise tracking.

\textbf{Sequence-Level Tracking:}
Graph-based approaches have gained popularity for maintaining long-term consistency. Brasó et al.~\cite{braso2020learning} and Cetintas et al.~\cite{cetintas2023unifying} frame association as edge prediction in a spatio-temporal graph, while Lu et al.~\cite{lu2024self} introduce self-supervised learning via path consistency. However, these still rely primarily on appearance embeddings rather than spatial-aware cues.

\textbf{Frame-Wise Tracking:}
Attention-based models have shown strong potential in object matching across frames. TrackFormer~\cite{meinhardt2022trackformer} and MOTRv2~\cite{zhang2023motrv2} leverage transformer decoders for joint detection and tracking, while MeMOTR~\cite{gao2023memotr} incorporates long-term memory. They perform well under non-linear motion datasets such as DanceTrack but poorly under linear motion datasets such as MOT17. Earlier dual-attention designs~\cite{zhu2018online} laid the foundation for such approaches. Regression-based methods like OC-SORT~\cite{cao2023observation}, Deep OC-SORT~\cite{maggiolino2023deep}, and DiffMOT~\cite{lv2024diffmot} emphasize motion continuity and efficient association. Confidence-aware~\cite{shim2024confidence} and post-correction strategies~\cite{huang2024deconfusetrack} have also emerged to reduce matching errors.

\noindent \textbf{Depth-Aware and Self-Supervised Association:}
Several works have introduced depth into tracking. Quach et al.~\cite{quach2024depth} apply relative depth ordering in Kalman filters, Wang et al.~\cite{wang2024robust} combine stereo depth with pose estimation, and Liu et al.~\cite{liu2022det} propose a depth-aware tracker for indoor scenes. However, these methods treat depth as an auxiliary signal rather than a primary association cue. Self-supervised Re-ID learning has gained traction recently. Li et al.~\cite{li2025self} embed self-supervision into FairMOT-style pipelines using path consistency, improving generalization. Nonetheless, it depends on contrastive or clustering objectives and does not utilize fused 3D spatial information.

\section{Proposed approach}
\label{sec:METHOD}

\begin{figure}[t]
    \centering
    \includegraphics[width=1\linewidth]{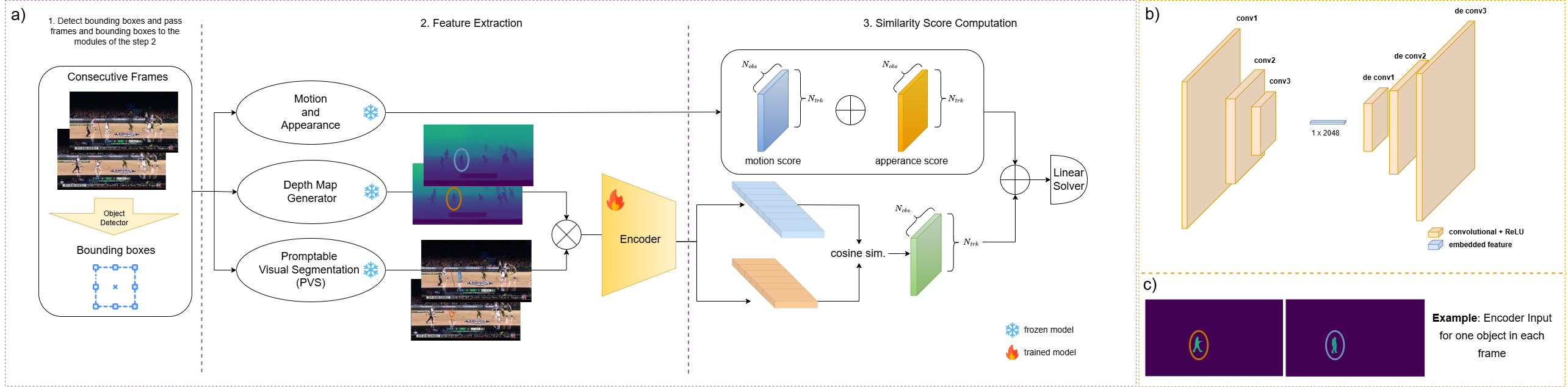}
    \caption{
    (a) Overview of SelfTrEncMOT. Given consecutive video frames and their object detector bounding boxes, we extract motion and appearance embeddings, and compute depth maps (via zero-shot monocular estimation) and segmentation masks (via Promptable Visual Segmentation). Depth and segmentation cues are fused into depth-segmentation embeddings and refined by a self-supervised encoder. The final association score integrates these embeddings with motion and appearance cues using a linear assignment solver. (b) Architecture of the depth-segmentation autoencoder. (c) Example of the encoder's input embedding.
    }
    \label{fig:spatio_depth_pipeline}
\end{figure}

Our approach follows the tracking-by-detection paradigm, where objects are first localized in each frame and then associated over time in a frame-wise manner. For detection, consistent with recent MOT benchmarks, we use YOLOX~\cite{zheng2021yolox}. As shown in Fig.~\ref{fig:spatio_depth_pipeline}, our method is composed of three main modules: a depth-segmentation fusion module that combines depth and mask features for each object (See Sec.~\ref{sec:depth_seg_fusion}); a self-supervised encoder that refines these fused features into stable embeddings (See Sec.~\ref{sec:spatio_depth_encoder}); and an appearance-motion module that extracts re-identification features and motion cues (See Sec.~\ref{sec:appearance_motion_fusion}).

Together, these modules generate three similarity scores that are used during object association. Following ByteTrack~\cite{zhang2022bytetrack}, the association process is divided into two stages. In the first stage, let \(\text{DC}\) denote the detection confidence score output by the object detector; high-confidence detections (\(DC > 0.6\)) are matched to existing tracklets using our multi-cue similarity scores (Sec.~\ref{sec:matching}). In the second stage, unmatched detections are associated with tracklets using Intersection-over-Union (IoU), based on predicted positions from a non-linear Kalman filter.

\subsection{Depth-Segmentation Fusion}
\label{sec:depth_seg_fusion}

\textbf{Zero-Shot Depth Estimation:}  
Each frame in the video sequence is processed using Depth Pro~\cite{bochkovskii2024depth} to generate a depth map, which provides a relative spatial representation of the scene.

\noindent \textbf{Promptable Visual Segmentation (PVS):}  
To achieve fine-grained shape alignment beyond bounding boxes, we incorporate PVS, a method that enables segmentation mask propagation across frames by leveraging prompts (e.g., points, bounding boxes, or masks). PVS extends static image segmentation to the video domain, facilitating consistent mask generation for the same object across time~\cite{delatolas2024learning, cheng2021modular, heo2020interactive}.

For implementation, we use Segment Anything Model 2 (SAM2)~\cite{ravi2024sam}, a prompt-driven segmentation framework designed for both images and videos. We use SAM2 for spatio-temporal shape alignment. Although SAM2 supports limited tracking, it is not designed for object ID consistency and bounding box matching as required in standard MOT tasks.

Our segmentation-driven fusion process proceeds as follows: For each tracklet in frame \(t{-}1\), the corresponding bounding box is used as a prompt input to SAM2 to generate a precise segmentation mask that reflects object shape within the box. Next, for each newly detected object in frame \(t\), its bounding box is used to prompt SAM2 in reverse (i.e., propagate backward) to recover its segmentation in frame \(t{-}1\), aligning it with existing tracklets for direct association.
Once segmentation masks are obtained for both existing tracklets (from \(t{-}1\)) and new detections (backward-propagated to \(t{-}1\)), we perform pixel-wise multiplication of each mask with its corresponding depth map. This yields fused depth-segmentation embeddings that encode both fine-grained object shape and relative spatial location. 

\subsection{Self-Supervised Depth-Segmentation Encoder}
\label{sec:spatio_depth_encoder}

Computing the mask IoU between segmentation outputs is computationally expensive and introduces major latency into the tracking pipeline. Additionally, both segmentation and depth estimation used in the depth-segmentation fusion are subject to limitations. SAM2 may produce misaligned masks when propagating segments across frames, particularly under fast motion or occlusion. We use DepthPro~\cite{bochkovskii2024depth}, a recent zero-shot monocular depth estimator, to obtain dense depth maps from RGB frames. DepthPro, while effective, can generate noisy depth maps in regions with poor texture or challenging lighting. These imperfections degrade the stability and reliability of the fused features.

To address this, we introduce a lightweight depth-segmentation encoder designed to suppress noise and enhance the temporal consistency of fused features. The encoder is part of the compact convolutional autoencoder that learns to denoise and compress the fused maps into discriminative embeddings suitable for tracking. The encoder uses three convolutional layers with kernel size $4 \times 4$ and stride 2, increasing channels from 1 to 32, 64, and 128, followed by batch normalization and ReLU activations. The resulting feature map is flattened and passed through a linear layer to produce a 2048-dimensional bottleneck. The decoder mirrors this with transposed convolutions that progressively upsample and reduce channels back to 1. The bottleneck embeddings, computed per object, serve as compact descriptors for matching, while reconstruction ensures that the bottleneck retains key structural cues. This encoder is trained in a self-supervised manner to enhance the discriminative quality and temporal consistency of the fused depth-segmentation features. The training process of our autoencoder proceeds as follows: At each training step, the depth-segmentation embeddings of tracklets from frame \(t{-}1\) and new detections from frame \(t\) are passed through a shared autoencoder. The training objective combines two terms: a reconstruction loss, computed as Mean Squared Error (MSE) between each input embedding \(f_i\) and its reconstruction \(\hat{f}_i\),
\begin{equation}
    \mathcal{L}_{\text{recon}} = \left\| f_i - \hat{f}_i \right\|_2^2,
\end{equation}
and a bottleneck consistency loss that enforces temporal alignment between embeddings at consecutive frames,
\begin{equation}
    \mathcal{L}_{\text{bottleneck}} = \left\| b_{t-1} - b_t \right\|_2^2,
\end{equation}
where \(b_{t-1}\) and \(b_t\) denote bottleneck features at frames \(t{-}1\) and \(t\). The final objective is the sum of both terms:
\begin{equation}
    \mathcal{L}_{\text{total}} = \mathcal{L}_{\text{recon}} + \mathcal{L}_{\text{bottleneck}}.
    \label{eq:totalloss}
\end{equation}

\begin{figure}[t]
    \centering
    \includegraphics[width=0.8\textwidth]{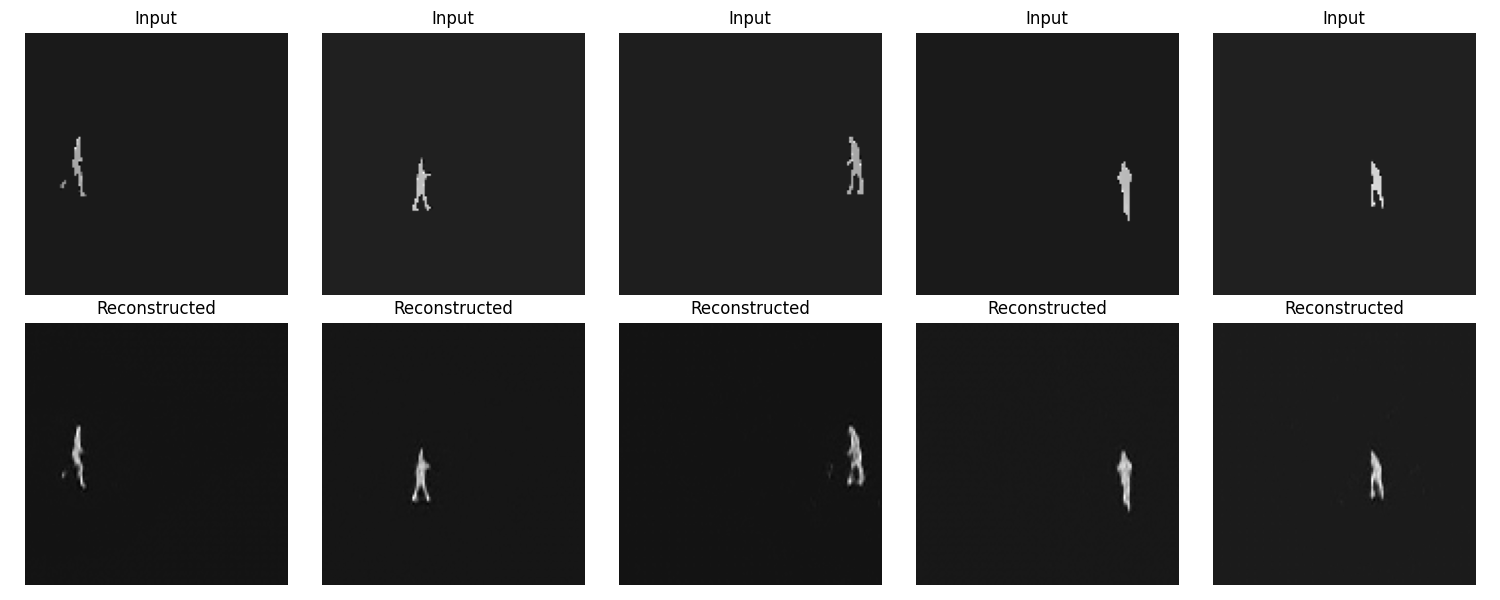}
    \caption{Qualitative results of the depth-segmentation autoencoder. Top: input fused embeddings; Bottom: reconstructions. The encoder preserves key spatial details and object boundaries, supporting robust association, despite variations in scale and structure.}
    \label{fig:autoencoder_recon}
\end{figure}

This self-supervised refinement not only filters out noise from the segmentation and depth sources but also improves temporal coherence in the feature embeddings, leading to more reliable object association under challenging visual conditions.  
To further stabilize temporal dynamics in the encoder features across frames, inspired by ~\cite{aharon2022bot, du2023strongsort}, we apply the tracklet embedding update strategy:
\begin{equation}
\label{eq:embupdate}
    emb_t = \mathcal{C} \cdot emb_{t-1} + (1 - \mathcal{C}) \cdot emb_{new},
\end{equation}

\noindent where \(\mathcal{C}\) is dynamically computed as in~\cite{maggiolino2023deep},  $\mathcal{C} = \mathcal{T} + (1 - \mathcal{T}) \cdot \left(1 - \frac{DC - \text{thresh}}{1 - \text{thresh}}\right)$, \(\text{DC}\) is the detection confidence score, \(\text{thresh}\) is a fixed confidence threshold (set to 0.6), and \(\mathcal{T}\) is a hyperparameter (set to 0.95). Figure~\ref{fig:autoencoder_recon} represents the input-output pairs of the depth-segmentation autoencoder, where the top row shows the input fused depth and mask-aware maps, and the bottom row illustrates the corresponding reconstructions. This visualization highlights how the encoder preserves critical structural cues despite compression. This result shows that despite the major reduction in dimension, the reconstructed masks retain useful information. The refined embeddings output by the encoder are then compared using cosine similarity to compute the depth-segmentation similarity score \(S_{sd}\). The raw depth-segmentation product is never directly used for matching; only the encoder-refined features participate in similarity computation.

\subsection{Appearance-Motion Fusion for MOT}
\label{sec:appearance_motion_fusion}

\textbf{Nonlinear Kalman Filter:}
Our SelfTrEncMOT method uses a nonlinear Kalman Filter to model object motion dynamics and predict tracklet locations across frames. This filter operates in two main steps: it first propagates previous state estimates based on learned motion priors, then corrects predictions using new observations. To enhance robustness to missed detections, we incorporate the observation-centric re-update (ORU) mechanism from OC-SORT~\cite{cao2023observation}, which interpolates virtual measurements when intermediate detections are absent. This approach allows tracklets to be updated more consistently over time, even during short-term occlusions. Each state includes bounding box geometry and motion velocities, enabling accurate position forecasting under complex dynamics.

\textbf{Motion Matching:}  
During association, a matching score matrix is computed using two motion-based components: \(S_{\text{IoU}}\) and \(S_{\text{ang}}\). \(S_{\text{IoU}}\) measures spatial overlap between predicted and observed bounding boxes, while \(S_{\text{ang}}\) captures angular consistency in motion direction. Higher scores in both terms reflect stronger association confidence~\cite{cao2023observation}.

\textbf{Appearance Matching:}  
We adopt FastReID~\cite{he2023fastreid}, a model based on convolutional neural networks (CNN) trained for MOT~\cite{maggiolino2023deep, lv2024diffmot}, to extract the appearance embeddings from each detected object. These features are compared across frames using cosine similarity to compute an appearance-based score matrix \(S_{\text{emb}}\).

\subsection{Total Matching Score and Linear Solver}
\label{sec:matching}

To associate detected bounding boxes with existing tracklets, we compute a total matching score that integrates motion, appearance, and depth-segmentation cues as follows:
\begin{equation}
    \label{eq:matchscore}
    \begin{aligned}
        Match_{t} = & \; S_{\text{IoU}_{t}}(\hat{X}, D) + S_{\text{ang}_{t}}(\hat{X}, D) + S_{\text{sd}_{t}}(\hat{X}, D) + S_{\text{emb}_{t}}(\hat{X}, D),
    \end{aligned}
\end{equation}
where \(\hat{X}\) denotes the predicted tracklets, \(D\) the detected bounding boxes, \(S_{\text{IoU}_{t}}(\hat{X}, D)\) denotes spatial overlap between predicted and observed bounding boxes, \(S_{\text{ang}_{t}}(\hat{X}, D)\) captures angular motion similarity~\cite{maggiolino2023deep}, \(S_{\text{emb}_{t}}(\hat{X}, D)\) measures appearance similarity using cosine distance between FastReID embeddings, and \(S_{\text{sd}_{t}}(\hat{X}, D)\) is the proposed depth-segmentation similarity score obtained from the cosine similarity between encoder embeddings. The matching score matrix \(Match_t\) is then negated to form a cost matrix suitable for linear assignment. We use the Hungarian algorithm~\cite{bewley2016simple} to perform optimal data association at each time step.

\section{Experimental Results}

\subsection{Datasets and Evaluation Metrics}
\label{sec:datasets_metrics}

\textbf{Datasets.}  
We evaluate our approach across three benchmarks to assess its effectiveness under varying motion dynamics, crowd densities, and interaction complexities.

\textit{SportsMOT} features fast, unpredictable subject movement, common in athletic or competitive contexts. The motion is dynamic, and inter-object occlusions occur frequently.

\textit{DanceTrack} presents performance-driven scenarios where subjects exhibit highly non-linear motion, frequent occlusions, and close-proximity interactions. The dataset includes 40 training, 25 validation, and 35 test sequences, placing emphasis on appearance similarity.

\textit{MOT17} is a widely used benchmark for multi-pedestrian tracking in urban environments, featuring scenes of moderate crowd density and structured pedestrian movement. It serves as a standard for evaluating trackers under relatively linear and predictable motion. While useful for general-purpose benchmarking, MOT17 is less representative of the non-linear and appearance-ambiguous motion patterns targeted by our method.

\noindent \textbf{Evaluation Metrics.}  
For a comprehensive assessment, we adopt standard MOT evaluation metrics: Higher Order Tracking Accuracy (HOTA), Association Accuracy (AssA), Detection Accuracy (DetA)~\cite{luiten2021hota}, Identification F1 Score (IDF1)~\cite{ristani2016performance}, and Multi-Object Tracking Accuracy (MOTA)~\cite{bernardin2008evaluating}. HOTA provides a balanced evaluation of detection and association accuracy. IDF1 and AssA specifically emphasize identity preservation and association quality, while MOTA focuses on detection-level performance. In terms of computational efficiency, we report frames per second (FPS) based on the tracking component, consistent with prior MOT works where runtime typically reflects the association stage~\cite{cao2023observation}.

\noindent \textbf{Implementation Details.}  
For consistency and fair benchmarking, we adopt YOLOX~\cite{zheng2021yolox} as the default object detector, following recent MOT methods~\cite{lv2024diffmot, cao2023observation, maggiolino2023deep}. In our tables, methods using YOLOX are highlighted in blue. 
We adhere to the standard tracking hyperparameters and evaluation settings (e.g., detection thresholds, update smoothing) from~\cite{maggiolino2023deep, shim2024confidence}. 

\noindent \textbf{Training.} SAM2 and DepthPro are not fine-tuned but used in their released form. The depth-segmentation autoencoder is trained offline, independently of the tracking method. Training is performed separately on fused depth-segmentation embeddings extracted from the training split of each dataset. We use a single NVIDIA A100 GPU with a batch size of 128 for all benchmarks. The method is trained for 12 epochs using the Adam optimizer with a learning rate of 1e\textsuperscript{-3}. The loss function in Eq.~\ref{eq:totalloss} combines two Mean Squared Error (MSE) terms: one for reconstruction and another for bottleneck alignment across frames, as described in Section~\ref{sec:spatio_depth_encoder}. To encourage stronger temporal consistency, only the bottleneck loss is used for the final two epochs. After training, only the encoder is retained and integrated into the tracker for inference. We train our method for each dataset separately.

\noindent \textbf{Inference.} All experiments are conducted on a single NVIDIA A100 GPU with a batch size of one. On the DanceTrack validation set, our tracker runs with the association stage sustaining over 125 frames per second (FPS). In our results tables, we divide MOT methods into TBD (bottom part) and JDR (upper part), and mark in bold the best-performing method in each category.

\noindent \textbf{Results for SportsMOT:}  
The SportsMOT benchmark poses unique challenges due to its dynamic trajectories and interactions between visually similar targets. As shown in Table~\ref{tab:sportmot_comparison}, compared to the state-of-the-art DiffMOT~\cite{lv2024diffmot}, our SelfTrEncMOT achieves better results across HOTA, AssA, and IDF1 association and identification metrics. These improvements highlight the effectiveness of depth-segmentation cues and our self-supervised encoder. 
\begin{table}[!ht]
  \centering
  \footnotesize
  \setlength{\tabcolsep}{3pt}
  \renewcommand{\arraystretch}{1.1}
  \resizebox{0.6\columnwidth}{!}{%
    \begin{tabular}{@{}lccc|cc@{}}
      \toprule
      Method & HOTA$\uparrow$ & IDF1$\uparrow$ & AssA$\uparrow$ & MOTA$\uparrow$ & DetA$\uparrow$ \\
      \midrule
      FairMOT~\cite{zhang2021fairmot} & 49.3 & 53.5 & 34.7 & 86.4 & 70.2 \\
      CenterTrack~\cite{zhou2020tracking} & 62.7 & 60.0 & 48.0 & 90.8 & 81.7 \\
      TransTrack~\cite{sun2020transtrack} & \textbf{68.9} & \textbf{71.5} & \textbf{57.5} & \textbf{92.6} & \textbf{82.7} \\
      \hline
      \hline
      \rowcolor{lightblue} ByteTrack~\cite{zhang2022bytetrack} & 62.8 & 69.8 & 51.2 & 94.1 & 77.1 \\
      \rowcolor{lightblue} BoT-SORT~\cite{aharon2022bot} & 68.7 & 70.0 & 55.9 & 94.5 & 84.4 \\
      \rowcolor{lightblue} OC-SORT~\cite{cao2023observation} & 71.9 & 72.2 & 59.8 & 94.5 & 86.4 \\
      \rowcolor{lightblue} DiffMOT~\cite{lv2024diffmot} & 72.1 & 72.8 & 60.5 & 94.5 & 86.0 \\
      \rowcolor{lightblue} *ByteTrack~\cite{zhang2022bytetrack} & 64.1 & 71.4 & 52.3 & 95.9 & 78.5 \\
      \rowcolor{lightblue} *MixSort-Byte~\cite{cui2023sportsmot} & 65.7 & 74.4 & 58.4 & 96.2 & 78.8 \\
      \rowcolor{lightblue} *OC-SORT~\cite{cao2023observation} & 73.7 & 74.0 & 61.5 & 96.5 & 88.5 \\
      \rowcolor{lightblue} *MixSort-OC~\cite{cui2023sportsmot} & 74.1 & 74.4 & 62.0 & 96.5 & 88.5 \\
      \rowcolor{lightblue} *GeneralTrack~\cite{qin2024towards} & 74.1 & 76.4 & 61.7 & 96.8 & 89.0 \\
      \rowcolor{lightblue} *DiffMOT~\cite{lv2024diffmot} & 76.2 & 76.1 & 65.1 & \textbf{97.1} & \textbf{89.3} \\
      \rowcolor{lightblue} *SelfTrEncMOT & \textbf{76.4} & \textbf{77.1} & \textbf{66.0} & 95.84 & 88.4 \\
      \bottomrule
    \end{tabular}%
  }
  \caption{
    Comparison with MOT trackers on the SportsMOT \textbf{test set}. The lower part shows TBD methods, which are relevant to ours. Methods marked with * indicate the detector was trained on SportsMOT's train and validation sets. YOLOX-based methods are highlighted in blue. The top part includes JDR methods, which have higher computational requirements for training than TBD methods.
  }
  \label{tab:sportmot_comparison}
\end{table}

\noindent \textbf{Results for DanceTrack:}  
We evaluate SelfTrEncMOT on the DanceTrack benchmark in Table~\ref{tab:dancetrack_comparison}. Compared to the TBD state-of-the-art, such as DiffMOT~\cite{lv2024diffmot}, Deep OC-SORT~\cite{maggiolino2023deep}, and CMTrack~\cite{shim2024confidence}, our SelfTrEncMOT achieves better performance in the association metrics HOTA, IDF1, and AssA.
The current state-of-the-art under this dataset is MOTRv2~\cite{zhang2023motrv2}, which is a Joint Detection-ReID, relying on a transformer-based architecture with joint training for detection and tracking. MOTRv2 requires higher computational requirements than our method, e.g., 8x Tesla V100 GPUs for training. Our method requires 1x GPU.
MOTRv2 underperforms for the linear motion dataset MOT17 (see Table \ref{tab:mot17_results}).

\begin{table}[!ht]
  \centering
  \setlength{\tabcolsep}{5pt}
  \renewcommand{\arraystretch}{1.2}
  \resizebox{0.6\columnwidth}{!}{%
    \begin{tabular}{@{}lccc|cc@{}}
      \toprule
      Method & HOTA$\uparrow$ & IDF1$\uparrow$ & AssA$\uparrow$ & MOTA$\uparrow$ & DetA$\uparrow$ \\
      \midrule
      FairMOT~\cite{zhang2021fairmot} & 39.7 & 40.8 & 23.8 & 82.2 & 66.7 \\
      CenterTrack~\cite{zhou2020tracking} & 41.8 & 35.7 & 22.6 & 86.8 & 78.1 \\
      TransTrack~\cite{sun2020transtrack} & 45.5 & 45.2 & 27.5 & 85.4 & 75.9 \\
      \rowcolor{lightblue} MOTRv2~\cite{zhang2023motrv2} & \textbf{69.9} & \textbf{71.7} & \textbf{59.0} & \textbf{91.9} & \textbf{83.0} \\
      \hline  
      \hline  
      \rowcolor{lightblue} ByteTrack~\cite{zhang2022bytetrack} & 47.3 & 52.5 & 31.4 & 89.5 & 71.6 \\
      \rowcolor{lightblue} MotionTrack~\cite{xiao2024motiontrack} & 52.9 & 53.8 & 34.7 & 91.3 & 80.9 \\
      \rowcolor{lightblue} OC-SORT~\cite{cao2023observation} & 55.1 & 54.2 & 38.0 & 89.4 & 80.3 \\
      \rowcolor{lightblue} StrongSORT++~\cite{du2023strongsort} & 55.6 & 55.2 & 38.6 & 91.1 & 80.7 \\
      \rowcolor{lightblue} GeneralTrack~\cite{qin2024towards} & 59.2 & 59.7 & 42.8 & 91.8 & 82.0 \\
      \rowcolor{lightblue} C-BIoU~\cite{yang2023hard} & 60.6 & 61.6 & 45.4 & 91.8 & 81.3 \\
      \rowcolor{lightblue} Deep OC-SORT~\cite{maggiolino2023deep} & 61.3 & 61.5 & 45.8 & 92.3 & 82.2 \\
      \rowcolor{lightblue} CMTrack~\cite{shim2024confidence} & 61.8 & 63.3 & 46.4 & 92.5 & - \\
      \rowcolor{lightblue} DiffMOT~\cite{lv2024diffmot} & 62.3 & 63.0 & 47.2 & \textbf{92.8} & \textbf{82.5} \\
      \rowcolor{lightblue} \textbf{SelfTrEncMOT} & \textbf{64.14} & \textbf{66.47} & \textbf{50.85} & 90.08 & 81.06 \\
      \bottomrule
    \end{tabular}%
  }
  \caption{
  Comparison on the DanceTrack \textbf{test set}. Methods are grouped into JDR in the upper part and TBD in the lower part. Methods using YOLOX are highlighted in blue. SelfTrEncMOT ranks first among TBD methods in identity association metrics (HOTA, AssA, IDF1). The state-of-the-art MOTRv2 requires more computational resources than our method (e.g., 8x GPU versus 1x GPU for training).
}

  \label{tab:dancetrack_comparison}
\end{table}

\noindent \textbf{Results for MOT17:}
As shown in Table~\ref{tab:mot17_results}, our method does not surpass state-of-the-art performance on MOT17. This is primarily because object motions in this dataset are mostly linear (e.g., cars and pedestrians moving in predictable directions). In such scenarios, motion patterns leave limited scope for depth information to provide substantial additional benefits. Nevertheless, SelfTrEncMOT achieves competitive results, demonstrating robust tracking performance comparable to leading methods.

\begin{table}[!ht]
  \centering
  \footnotesize
  \setlength{\tabcolsep}{2pt}
  \renewcommand{\arraystretch}{1.1}
  \resizebox{0.8\columnwidth}{!}{%
  \begin{tabular}{@{}lccc|c|ccccc@{}}
    \toprule
    \textbf{Method} & HOTA$\uparrow$ & IDF1$\uparrow$ & AssA$\uparrow$ & MOTA$\uparrow$ & FP($10^4$)$\downarrow$ & FN($10^4$)$\downarrow$ & ID$_s$$\downarrow$ & Frag$\downarrow$ & AssR$\uparrow$ \\
    \midrule
    FairMOT~\cite{zhang2021fairmot} & 59.3 & 72.3 & 58.0 & 73.7 & 2.75 & 11.7 & 3,303 & 8,073 & 63.6 \\
    CenterTrack~\cite{zhou2020tracking} & 52.2 & 64.7 & - & 67.8 & 1.8 & 1.6 & 3,039 & - & - \\
    TransTrack~\cite{sun2020transtrack} & 54.1 & 63.5 & 47.9 & 75.2 & 5.02 & 8.64 & 3,603 & 4,872 & 57.1 \\
     \rowcolor{lightblue} MOTRv2~\cite{zhang2023motrv2} & \textbf{62.0} & \textbf{75.0} & \textbf{60.6} & \textbf{78.6} & - & - & - & - & - \\
    \hline
    \hline
    \rowcolor{lightblue} ByteTrack~\cite{zhang2022bytetrack} & 63.1 & 77.3 & 62.0 & 80.3 & 2.55 & 8.37 & 2,196 & 2,277 & 68.2 \\
    \rowcolor{lightblue} MotionTrack~\cite{xiao2024motiontrack} & 65.1 & 80.1 & - & \textbf{81.1} & 2.38 & \textbf{8.16} & 1,140 & - & - \\
    \rowcolor{lightblue} OC-SORT~\cite{cao2023observation} & 63.2 & 77.5 & 63.2 & 78.0 & \textbf{1.51} & 10.8 & 1,950 & 2,040 & 67.5 \\
    \rowcolor{lightblue} StrongSORT++~\cite{du2023strongsort} & 64.4 & 79.5 & 64.4 & 79.6 & 2.79 & 8.62 & 1,194 & 1,866 & \textbf{71.0} \\
    \rowcolor{lightblue} GeneralTrack~\cite{qin2024towards} & 64.0 & 78.3 & 63.1 & 80.6 & - & - & 1,563 & - & - \\
    \rowcolor{lightblue} C-BIoU~\cite{yang2023hard} & 64.1 & 79.7 & 63.7 & \textbf{81.1} & - & - & - & - & - \\
    \rowcolor{lightblue} Deep OC-SORT~\cite{maggiolino2023deep} & 64.9 & 80.6 & 65.9 & 79.4 & 1.66 & 9.88 & 1,023 & 2,196 & 70.1 \\
    \rowcolor{lightblue} CMTrack~\cite{shim2024confidence} & \textbf{65.5} & \textbf{81.5} & \textbf{66.1} & 80.7 & 2.59 & 8.19 & \textbf{912} & \textbf{1,653} & - \\
    \rowcolor{lightblue} DiffMOT~\cite{lv2024diffmot} & 64.5 & 79.3 & 64.6 & 79.8 & - & - & - & - & - \\
    \rowcolor{lightblue} SelfTrEncMOT & 63.48 & 78.12 & 63.25 & 79.16 & 1.9 & 9.6 & 1,008 & - & 68.8 \\
    \bottomrule
  \end{tabular}%
  }
  \caption{Comparison on the MOT17 \textbf{test set}. The JDR methods are in the upper part, and the TBD methods are in the lower part. Methods using YOLOX are highlighted in blue. State-of-the-art is CMTrack, which performs weakly for challenging datasets such as DanceTrack. MOTRv2 ranks lower than all TBD methods, including ours.}
  \label{tab:mot17_results}
\end{table}

\subsection{Ablation Study}

To assess the effectiveness of depth-segmentation guided association in our method, we perform an ablation study (Table~\ref{tab:ablation_depth_encoder}) comparing three setups: (1) appearance with mask IoU, (2) appearance with bounding box IoU~\cite{maggiolino2023deep}, and (3) appearance with bounding box IoU augmented by our depth-segmentation encoder. We evaluate on the \textit{DanceTrack} and \textit{MOT17} validation sets, which represent non-linear and structured pedestrian motion, respectively.
\begin{table}[!ht]
  \centering
  \resizebox{0.8\columnwidth}{!}{%
  \begin{tabular}{@{}cccc|ccc|ccc@{}}
    \toprule
    \multicolumn{4}{c|}{\textbf{Ablation Setting}} & \multicolumn{3}{c|}{\textbf{DanceTrack-val}} & \multicolumn{3}{c}{\textbf{MOT17-val}} \\
    \midrule
    Appearance & Mask IoU & Bbox IoU & Depth-Segmentation & HOTA$\uparrow$ & AssA$\uparrow$ & IDF1$\uparrow$ & HOTA$\uparrow$ & AssA$\uparrow$ & IDF1$\uparrow$ \\
    \midrule
    \checkmark & \checkmark &  &  & 54.78 & 38.52 & 52.71 & 68.26 & 66.81 & 77.20 \\
    \checkmark &  & \checkmark &  & 59.46 & 43.93 & 59.11 & 70.43 & 70.83 & 80.73 \\
    \checkmark &  & \checkmark & \checkmark & \textbf{60.61} & \textbf{47.04} & \textbf{62.34} & \textbf{72.22} & \textbf{71.79} & \textbf{82.52} \\
    \bottomrule
  \end{tabular}%
  }
  \caption{Ablation study on DanceTrack and MOT17 \textbf{validation sets}. We evaluate combinations of appearance and IoU-based cues, with and without our encoder-refined depth-segmentation association module.}
  \label{tab:ablation_depth_encoder}
\end{table}

\noindent \textbf{Complementarity with Box-Level Cues:}  
Switching from mask IoU to bounding box IoU improves performance significantly, particularly in structured settings like MOT17. However, further gains from depth-segmentation integration indicate that relying on 2D overlap alone remains insufficient for robust identity tracking, especially under occlusion.

\noindent \textbf{Role of Depth-Segmentation Association:}  
Adding the encoder-refined depth-segmentation embeddings to appearance and bounding box cues yields consistent gains across datasets. The improvement is especially pronounced on DanceTrack, where motion is more irregular and visual ambiguity is higher, highlighting their complementary value beyond 2D geometry.

\section{Conclusion}
We presented SelfTrEncMOT, a novel tracking method that incorporates encoder-refined depth-segmentation embeddings as a standalone cue for robust association. By combining zero-shot monocular depth with promptable segmentation and refining the fused cues via self-supervised learning, our approach captures fine-grained spatial and geometric context often missed by traditional motion or appearance signals. Evaluations on diverse benchmarks demonstrate strong identity preservation under occlusion, ambiguity, and crowd density, underscoring the value of pixel-aligned, depth-aware reasoning for MOT.

While the association module is efficient, the overall runtime is limited by the depth estimation step: DepthPro~\cite{bochkovskii2024depth} requires about 0.3 seconds per frame, as reported in its original paper, and forms the primary bottleneck. Future work will investigate real-time depth estimators to improve speed without degrading accuracy, as well as contrastive objectives for the encoder to enhance the discriminability and robustness of depth-segmentation embeddings.

\clearpage
\bibliography{egbib}
\end{document}